\documentclass[10pt,twocolumn,letterpaper]{article}
\usepackage{times}
\usepackage{epsfig}
\usepackage{amsmath, amssymb, graphicx}
\usepackage{multirow}
\usepackage[dvipsnames]{xcolor}
\usepackage{authblk}
\usepackage[margin=1in]{geometry}
\usepackage[square,sort,comma,numbers]{natbib}
\definecolor{mygreen}{rgb}{0,0.7,0}
\usepackage{color, soul}

\begin{document}
\date{\vspace{-5ex}}
\title{Semi-Supervised Domain Adaptation with Representation Learning for
Semantic Segmentation across Time}

\author[1,2]{Assia Benbihi}
\author[3]{Matthieu Geist}
\author[1,4]{C{\'e}dric Pradalier}
\affil[1]{UMI2958 GeorgiaTech-CNRS}
\affil[2]{Centrale Sup{\'e}lec, Universit{\'e} Paris Saclay, Metz}
\affil[3]{Google Research, Brain Team}
\affil[4]{GeorgiaTech Lorraine}

\maketitle
\begin{abstract}

  Deep learning generates state-of-the-art semantic segmentation
  provided that a large number of images together with pixel-wise annotations
  are available. 
  To alleviate the expensive data collection process, 
  we propose a semi-supervised domain adaptation method for the
  specific case of images with similar semantic content but different pixel
  distributions. %
  A network trained with supervision on a past dataset
  is finetuned on the new dataset to conserve its features maps.
  The domain adaptation becomes a simple regression between feature
  maps and does not require annotations on the new dataset. %
  This method reaches performances similar to classic transfer
  learning on the PASCAL VOC dataset with synthetic transformations.

\end{abstract}

\section{Introduction}

\begin{figure}[htb]
  \centering
      \includegraphics[width=\linewidth]{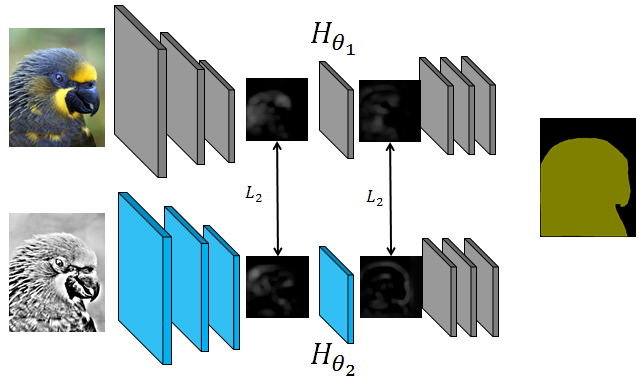}
  \caption{Top: The trained and frozen (gray) network provides ground truth deep representations. 
  Down: The trainable layers (blue) must learn the deep representations.}
  \label{fig:method}
\end{figure}

Recent deep learning applications such as environment monitoring
\cite{richard2018} and autonomous driving \cite{cordts2016cityscapes} rely on
semantic segmentation of datasets with redundant images. As monitoring
progresses and new datasets are acquired, their data distributions may change
due to light variations or camera upgrades. This can prevent a Convolutional
Neural Network (CNN) trained at time $t$ to generalize at a later time. Each
time the distribution deviates from the past one, the network needs to be
fine-tuned or retrained from scratch, both of which require ground-truth
pixel-wise annotations. To avoid the burden of annotating new datasets, we
propose a semi-supervised method to transfer the network knowledge across time.
We do so by transferring the deep representations learned by the network
instead of classic transductive transfer learning \cite{pan2010survey}, also
called finetuning. Figure \ref{fig:method} illustrates our method on a
synthetic transformation emulating a camera downgrade: a first network
$H_{\theta_1}$ is trained on a dataset $D_1$ with pixel-wise supervision. Then,
a second network $H_{\theta_2}$ is trained to generate deep representations for
a different dataset $D_2$ to match the deep representations of
$H_{\theta_1}$ on $D_1$. 
This approach has been studied in
\cite{bengio2012deep} for classification in the Unsupervised and Transfer
Learning Challenge. One of the limitations in this work was the lack of
instances pairs with the same semantic content. In the segmentation
applications we address, such pairs arise naturally which solves this issue.%

One application example is environmental monitoring across time
\cite{richard2018}: a DCNN is used to segment land categories on aerial images
over a period from 1955 to 2015. The 2015 dataset is made of RGB digital images
whereas the 1955 data is made of black and white analog images that have been
digitized. 
A network trained on the 2015 data can
not generalize on the 1955 one because the images do not have the same color domain
nor the same resolution or grain. And even though the surveys cover the same
area, the images are not perfectly aligned due to small changes in land use,
which prevent the use of 2015 annotations to train a network on the 1955 data.
What stays coarsely invariant for one image across surveys is the high-level
semantic content.

We emulate equivalent transformations on the PASCAL VOC 12 dataset~\cite{everingham2015pascal}.
Synthetic data is necessary to gather ideal
baselines since they require pixel-wise annotated images over several input
distribution (Section 3.2). To the best of our knowledge, there is no such
dataset publicly available. Experiments show that our transfer method reaches
the same performance as standard transfer learning and classic training. Our
contribution is twofold: i) we extend the initial work of \cite{bengio2012deep}
to semantic segmentation on a wider database, ii) we provide visual insight
on the CNN adaptation behavior. The next section describes the state-of-the-art
in transfer learning for DCNN. Section 3 and 4 present our method and
results.%

\section{Related work}

Domain adaptation through feature transfer is first applied to
classification: \cite{glorot2011domain}~uses auto-encoders to learn high-level
representations of sentiments from various datasets to better classify Amazon
reviews. And \cite{bengio2012deep} tackles the challenge of digits
classification where some classes are not represented in the training set.
Both learn a data transformation that best captures each class's unique features
so that a trained classifier can even process class instances from new domains.
This transformation is evaluated with the classification score. One of the best
methods relies on unsupervised auto-encoder training that forces the network to
capture the relevant features to reconstruct the input. 
 
We draw inspiration from these works and transfer the features maps of a
segmentation CNN across data domains. Our work differs from
\cite{glorot2011domain,bengio2012deep} in the target task and the feature
learning method. First, contrary to previous classification focus, we test our
method on semantic segmentation on the same dataset across time. Also, we tackle
a different category of domain adaptation where the output variable is the same
but the input pixel distribution varies. Second, we do not follow the previous
feature training methods: we use the annotations from one dataset to train a
segmentation network, then utilize the features that the network autonomously
learned as ground-truth feature representations. For every new segmentation
dataset, we force the network to match these representations. This method is
semi-supervised since it requires only to match images of the same locations,
which can be done during the data collection process using odometry.

Given that \cite{richard2018} did not release their segmentation dataset and
\cite{griffith2017symphony} only provides odometry datasets, we rely on three 
synthetic transformations of the PASCAL VOC 12 dataset \cite{everingham2015pascal}
to test our method.
Our transfer learning method reach the same
performance as finetuning and standard training. %

\section{Method}

\subsection{Training}

Let $H_{\theta}$ be a network model with weights $\theta$, $X$ an image and 
${\cal F}_{\theta}(X)$ the set a set of feature maps of the network $H_{\theta}$,
with ${\cal F}_{\theta}(X) = \left\{F^l_{\theta}(X), \;l \in L\right\}$.
$H_{\theta_1}$ is an instance of $H$ trained with supervision on $D_1$.
$H_{\theta_2}$ is another instance of $H$ first initialized with the weights
$\theta_1$. $H_{\theta_2}$ is then finetuned on image pairs $\left\{(X_1,
X_2)\subset D_1 \times D_2\right\}$ with the same semantic content but
different pixel distributions. The feature maps ${\cal F}_{\theta_2}(X_2)$
must match the corresponding ${\cal F}_{\theta_1}(X_1)$ without ground-truth
annotations on $D_2$. A loss is computed for each feature map and is
back-propagated only in the layers lower than $l$: $ \mathcal{L}^l(X) = w_l
\| F^l_{\theta_1} - F^l_{\theta_2} \|^2, w_l \in \mathbb{R}$. In our
implementation, the feature weights $w_l$ are constant over the training.
Investigations over dynamic weighting strategies have been left for future
work.

\subsection{Evaluation}

Our transfer method is evaluated with the segmentation performance of the adapted
network $H_{\theta_2}$ on the new dataset $D_2$. We use the standard
segmentation evaluataion metrics: the mean accuracy (acc) and the mIOU.
We compare against three baselines:
baseline $B_0$ measures the performance of the first
network $H_{\theta_1}$ on the new dataset $D_2$, i.e. how well $H_{\theta_1}$
generalizes to a dataset with the same content but a different pixel
distribution. This also gives a quantitative measure of the image
transformation between two datasets.%
In $B_1$, we train
$H_{\theta_2}$ with full supervision on $D_2$ using the pixel-wise annotations
from $D_1$. 
This ideal setting provides the performance our transfer learning should aim
at.  The last baseline $B_2$ measures the performance of $H_{\theta_2}$ when it
is initialized with $\theta_1$ and then classically fine-tuned on $D_2$ with
$D_1$ annotations.  This provides the performance of classic supervised
fine-tuning our method should reach while not using explicit annotations.

\subsection{Visualization}
Besides the proposed domain adaptation techniques described above, this
paper also developed a visualization technique inspired from existing
approaches used in other contexts.  Specifically, we design an optimization to
observe the evolution of the representations induced by the feature regression during
the domain adaptation. 
An image $X_1 \in D_1$ is fed to $H_{\theta_2}$ to generates a set of 
features ${\cal F}_{\theta_2}(X_1)$. Starting from white noise, we then generate 
the image $\hat{X_1}$ that minimizes 
$\sum_{l \in L} \|F^l_{\theta_2}(X_1) - F^l_{\theta_2}(\hat{X_1})\|$, i.e. a
version of $X_1$ as seen by $H_{\theta_2}$.
Visually, $\hat{X}_1$ has the same content as $X_1$ with the visual
aspect of $D_2$.

The optimization is inspired from the feature map inversion method
\cite{Mahendran_2015_CVPR} to integrate style transfer into it. In
addition, we adapt the style transfer method from \cite{gatys2016image}
\cite{johnson2016perceptual} to generate image content and style from only one
network instead of two. 
The final optimization is performed as follows: we initialize $\hat{X_1}$ with
white noise and feed it to $H_{\theta_2}$. For each feature map, a content loss and 
a style loss are computed and backpropagated into the image. 
$\hat{X_1}$ is optimized with Stochastic Gradient Descent (SGD). 
We use the losses and Gram matrix definitions
from \cite{gatys2016image}. The content loss is a simple $\mathcal{L}_2$ loss
between feature maps $\mathcal{L}_{content}(l)  = \frac{1}{2} \|F^l_{\theta_2}(X_1) - F^l_{\theta_2}(X)\|^2 $.
The style loss is a normalized $\mathcal{L}_2$ loss 
between the Gram matrix of the feature maps. This matrix is computed from a 2D
projection of each feature map $F^l_{\theta_2}$: 
$\mathcal{L}_{style}(l) \propto \|G(F^l_{\theta_2}(X_1))-G(F^l_{\theta_2}(X))\|^2 $.

\section{Experiments}

\subsection{Dataset}

We run tests on three synthetic transformations of the augmented version 
\cite{hariharan2011semantic} of the PASCAL VOC12 dataset
\cite{everingham2015pascal} resulting in 10 582 training images and 1449
validation images with 21 semantic classes.
The regression is trained on the 10 582 original images of $D_1$ and the  
transformed ones of $D_2$. $H_{\theta_2}$ is then evaluated 
on the transformed validation set of $D_2$.

\begin{figure}[thb]
  \centering
  \includegraphics[width=\linewidth]{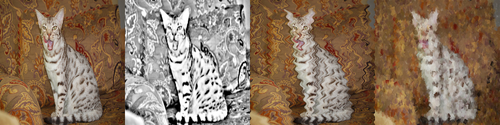}
  \caption{Synthetic transformations. Column 0: PASCAL. Left-Right: transformation.
   Photocopy (Distortion: 32.5\%), Ripple (62.6\%), Cubism (94.0\%)}
  \label{fig:transfoVOC12}
\end{figure}

Three transformations $T_1, T_2, T_3$ with increasing perturbations
are generated with GIMP \footnote{https://www.gimp.org/}
resulting in $D^1_2, D^2_2, D^3_2$ (Figure \ref{fig:transfoVOC12}). 
We use the 'photocopy' filter $T_1$ to emulate a change of color and
saturation. This problem arises in long-term environmental monitoring 
where recent datasets are numerical RGB images and older datasets are
collected with the numerization of analogic pictures \cite{richard2018}.
We address the issue of image misalignment and noise with the ripple distortion $T_2$.
This is typical in natural environment monitoring such as in the dataset from
\cite{griffith2017symphony}. 
Finally, we mix a change of texture and misalignment with edge noise in the 
cubism filter $T_3$.

For each transformation $T_i$, the image distortion between $D_1$ and $D_2^i$ 
is quantized 
with the normalized performance degradation of the network $H_{\theta_1}$ on 
$D_2^i$. After training $H_{\theta_1}$ on $D_1$ with the original 
DeepLab V3 \cite{chen2018deeplab} setup, the accuracy and mIOU reach
respectively 79.92\% and 69.22\%. 
With $\textrm{acc}(H_{\theta_1}, D_2)$ and $\textrm{mIOU}(H_{\theta_1}, D_2)$ 
the performances of $H_{\theta_1}$ on $D_2$, the image
distortion between $D_1$ and $D_2$ is:
$  \frac{1}{2} \left( \frac{|79.92 - \textrm{acc}(H_{\theta_1}, D_2)|}{79.92} + 
  \frac{|69.22 - \textrm{mIOU}(H_{\theta_1}, D_2)|}{69.22} \right)$.

The dataset distortions (Fig. \ref{fig:transfoVOC12}) comfort the visual intuition that
the three transformations exhibit an increasing level of complexity that
challenges the network trained only on $D_1$. 

\subsection{Experimental Setup}

\subsubsection{Supervised training of $H_{\theta_1}$.}
$H$ follows the state-of-the-art VGG-16 architecture \cite{simonyan2014very} 
from DeepLabV3 \cite{chen2018deeplab} for training time considerations. 
and converges within only 5 hours of training on 
an NVIDIA 1080Ti. As in \cite{chen2018deeplab},
we train the network for 20 000 iterations with a batch size of 10, SGD 
with a momentum of 0.9, a weight decay of 0.5 and the "poly"
learning rate policy initialized at $2.5 \times 10^{-4}$ and
$power=0.9$. 

\subsubsection{Feature map training of $H_{\theta_2}$.}
We run the unsupervised domain adaption training on several sets of  ${\cal F}$ 
to better understand the hierarchical model of network representations. 

An intuition gathered from the literature 
\cite{dosovitskiy2016inverting,oquab2014learning,simonyan2013deep} suggests that 
early layers capture low-level representations such as colors and edges,
whereas higher layers embed more complex representations such as object contours 
and their label. 
According to this intuition, we could expect that regression of high layers are
more relevant than lower ones when the transformation is significant.
To test this assumption, we run individual regressions on one feature map and compare 
the performance of $H_{\theta_2}$. We choose the feature map post max-pooling
as they show the highest representations shifts in consecutive layers. When we
display features of a VGG bloc, the features of successive convolutional layers
look highly similar whereas we always observe significant variations after the
pooling layers.

We also test the correlations between the network representation levels. 
We run the regression on all the post-pooling layers simultaneously with two
weighting strategies. 
In the first one, $W_{inc}$, $w_l$ increases with the layer level, i.e. we favor the deeper
representations. We do the contrary with the second one, $W_{dec}$, and rely more on
low-level features. In both cases, we use the following uniform weight distribution 
$[0.2, 0.4, 0.6, 0.8, 0.9]$. 

\subsection{Experimental Results}

\begin{figure}[thb]
  \includegraphics[width=\linewidth]{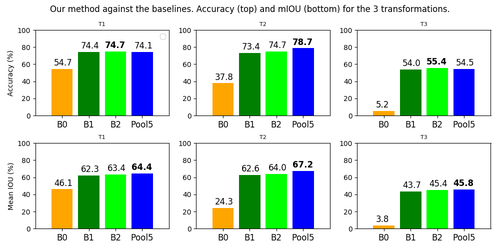}
  \caption{Transfer performance against the baselines.}
  \label{fig:metrics_baseline}
\end{figure}

\begin{figure}[thb]
  \includegraphics[width=\linewidth]{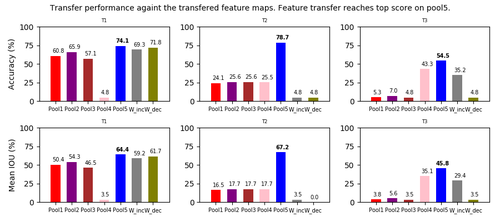}
  \caption{Transfer
  performance against the transferred features maps. Transferring on
  \texttt{pool5} gives the best scores.}
  \label{fig:metrics_layer}
\end{figure}

\begin{figure}[thb]
  \includegraphics[width=\linewidth]{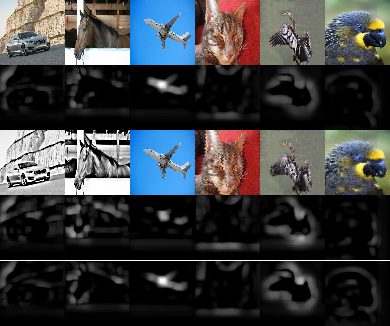}
  \caption{
    Deep representation evolution.
   Line 1: original images. 
   Line 2: one feature of the old network $H_{\theta_1}$ on the old data $D_1$. 
   Line 3: perturbed image from the new data $D_2$. 
   Line 4: the same feature of the old network $H_{\theta_1}$ on the new data $D_2$. 
   Line 5: the same feature of the new network $H_{\theta_2}$ on the new data $D_2$.
 }
  \label{fig:visu_feat}
\end{figure}

\begin{figure}[thb]
  \includegraphics[width=\linewidth]{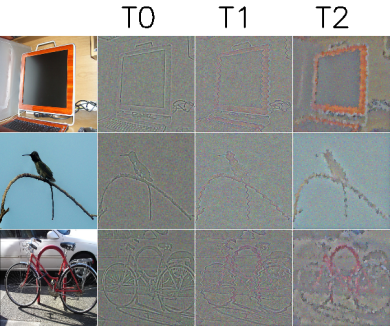}
  \caption{
    Style reconstruction. The finetuned network is fed with the left-most image. 
  We reconstruct the image as seen by the three finetuned networks (right
  images).
  The generated image has the same content and the style on which
  the network is finetuned.
}
  \label{fig:visu_style}
\end{figure}

Despite training without supervision, our method reaches similar or higher
performance than classic supervised fine-tuning
(Fig.~\ref{fig:metrics_baseline}). The feature map chosen for the regression
has a high impact on the performance: regression on the higher level maps
provides better results (Fig.~\ref{fig:metrics_layer}). Figure
\ref{fig:visu_feat} illustrates how the feature maps of the old network shift
toward their initial representations after our finetuning. The image
reconstructed from the finetuned feature maps displays the same content as in
the old dataset but with the style of the new one (Fig. \ref{fig:visu_style}).
This observation follows previous findings in neural image style transfer
\cite{gatys2016image}: $\mathcal{L}_2$-regression between feature maps does
transfer the image content but not the image style.

\paragraph{Quantitative Results}

Fig. \ref{fig:metrics_baseline} shows that the regression results on \texttt{pool5} reach similar 
performances to classic fine-tuning. 
The $B_0$ line recalls the performance of $H_{\theta_1}$ on the transformed dataset $D_2$. 
Our method improves significantly the metrics compared to $B_0$ which means that 
transferring deep representations is indeed a relevant transfer learning method for 
semantic segmentation.
Classic finetuning $B_2$ outperforms the training from scratch $B_1$ on $D_2$: this comforts
the results from \cite{lamblin2010important} on the importance of weight initialization 
and that finetuning is relevant for boosting the performances.
Our method always outperforms $B_2$'s mIOU and reaches similar mean accuracy. This shows that 
transfer learning of deep representations can replace classic CNN finetuning
without performance degradation.

Fig. \ref{fig:metrics_layer} summarizes the performances of individual and parallel regressions 
to better understand the network representation hierarchy.
The best performances are reached with the individual regression on the highest
post-pooling layer \texttt{pool5}.
This suggests that high-level representations are the most relevant to transfer for 
semantic segmentation. 
The metrics for each regression vary with the transformation type. 
The experiments do not allow to draw a general conclusion but we can gather an
intuition on which representations are the most relevant to transfer.
For the color and saturation transformation $T_1$, the transfer of layers up to 
{\tt pool3} improves $B_0$ which is not the case for $T_2$ and $T_3$. One explanation 
can be that $T_1$ conserves the alignment of the image and that color
processing is handled in the low-level layers.
$T_2$ and $T_3$ maintain the image color domain but modify the  
contours of the semantic units either with regular noise as in the ripple
transformation or with random one in the cubism effect. 

Object contours are features usually generated in deeper layers which 
can explain why the transfer of \texttt{pool\_5} only is relevant for these datasets.
These results also suggest that when low-level layers are not relevant to transfer, they
can hinder the transfer of the relevant layers. For example, the 
transfer learning on multiple layers performs worse than the transfer on \texttt{pool5} only. 
For $T_3$, we observe that a weight distribution that favors high layers performs
better than $W_{dec}$ but worse than the transfer of \texttt{pool5}. The
overall intuition we can draw consists in relying on high-level feature
transfer even though low-level layers can be relevant for color domain
perturbations. 

\paragraph{Qualitative Results}
Figure \ref{fig:visu_feat} illustrates how finetuning shifts the feature maps
on the new dataset towards the maps from the old one. The first two lines
display the original image and one of their feature maps. The next line
displays the perturbed images followed by the same feature map, before and after
our finetuning. The new map appears less noisy, especially
around the semantic contours of the image. This empirical observation
generalises to most of the image feature maps. One possible interpretation is
that our adaptation regularises the feature noise introduced by the change in 
image distribution.

The image reconstructed from the new CNN feature maps displays the visual style
of the new dataset (Fig. \ref{fig:visu_style}). This interesting observation
suggests that the $\mathcal{L}_2$ regression of feature maps only transfers
image content between CNNS and not the style. The new network is specialised on
the new image style. This explanation is coherent with previous image style
transfer work \cite{gatys2016image, johnson2016perceptual}: the optimisation
run in these work generally have two terms: the first one is a simple
$\mathcal{L}_2$ between feature maps and is called the `content' loss. The
second term, termed the `style loss', is the difference between the gram matrix
of the feature maps i.e. the relative relations of the vectors inside one
feature maps. Since we only regress the feature maps, it was expected
that the new network would embed the style of the new dataset.

\section{Conclusion}
We have introduced a method for unsupervised domain adaptation for semantic
segmentation that relies on the transfer of CNN deep representations. 
It is relevant for applications with redundant semantic content and a drift of
the pixel distributions such as autonomous-driving or long-term environment
monitoring for which new datasets covering a similar semantic content are acquired over time. 
This method shows similar performance to classic fine-tuning on three synthetic
transformations of the PASCAL VOC dataset that emulates color domain
variations, resolution degradation and noise.
Quantitative results suggest that high-level representations are the most relevant 
to transfer even though low-level transfer also reach acceptable performance
for color-domain transformations. These observations comfort the recurrent
intuitions on the semantics of CNN features.

\bibliographystyle{acm}
\bibliography{../samplepaper}

\end{document}